\documentclass{article}
\usepackage[utf8]{inputenc}
\usepackage[semicolon]{natbib}
\usepackage[pdftex]{graphicx}

\title{Data Uncertainty without Prediction Models}
\author{
  Bongjoon Park\\
  Samsung Display\\
  \texttt{bongjoon@dm.snu.ac.kr} \\
  \and
  Eunkyung Koh \\
  Samsung Display\\
  \texttt{eunkyung.koh@samsung.com } \\}
\date{October 2021}

\begin{document}

\maketitle
\begin{abstract}
Data acquisition processes for machine learning are often costly. 
To construct a high-performance prediction model with fewer data, 
a degree of difficulty in prediction is often deployed as the acquisition function in adding a new data point. 
The degree of difficulty is referred to as uncertainty in prediction models. 
We propose an uncertainty estimation method named a Distance-weighted Class Impurity without explicit use of prediction models. 
We estimated uncertainty using distances and class impurities around the location, and compared it with several methods based on prediction models for uncertainty estimation by active learning tasks. We verified that the Distance-weighted Class Impurity works effectively regardless of prediction models.
\end{abstract}

\section{Introduction}
Machine learning requires sufficient amount and quality of data. However, processes of data acquisition like labelling, measurements or simulations are often costly and time-consuming, and possibly involve destructive inspection. To reduce the cost of data acquisition, active learning that constructs prediction models with a small-sized initial data then selects data points with high uncertainty to be added into the learning data can be implemented. Uncertainty, here, is evaluated based on the initial prediction models. \citep{Cohn, McCallum}

Uncertainty is estimated based on the results of prediction models after learning models such as Support Vector Machine \citep{Platt99, Niaf, Patra} and Convolutional Neural Network (CNN) \citep{Gal_image, Wang, shen}. Because this estimated uncertainty is the result of prediction models, prediction model selection and training must take place prior to data acquisition processes. The model for predicting good results depends on the state of data, and the method of estimating uncertainty varies depending on the model. Thus, Uncertainty is uncertain in situations where prediction models are changed or train data are insufficient.

In this work, we propose a method for estimating uncertainty called a Distance-weighted Class Impurity in classification tasks of data without using prediction models. The proposed method measures distances from labelled data in the neighborhood, obtains the impurity of the class weighted in distance, and estimates the uncertainty by dividing the weighted impurity by the density of surrounding data. The Distance-weighted Class Impurity represents the probability that a decision boundary of real data exists at that location, and allows us to find the area that needs to be explored to pinpoint the decision boundary.

We tested the performance of the Distance-weighted Class Impurity with active learning on classification and regression tasks. We compared prediction accuracies of data collected through uncertainty based sampling using prediction models and Distance-weighted Class Impurity based sampling without using prediction models. 

We summarize our contributions as follows.
\begin{itemize}
\item We propose an estimating method of uncertainty, which does not require a machine learning model. 
\item We present an efficient data sampling method without specifying prediction models and checking the prediction accuracy. 
\end{itemize}

\section{DISTANCE-WEIGHTED CLASS IMPURITY}
In this section we define the Distance-weighted Class Impurity, which represents uncertainty of classes. We define where there are higher uncertainties in Section \ref{ch:uc_def}, propose a new metric of uncertainty named the Distance-weighted Class Impurity in Section \ref{ch:DCI}, and visualize it in Section \ref{ch:ex2d}.

\begin{figure}
\centering
\includegraphics[scale=0.38]{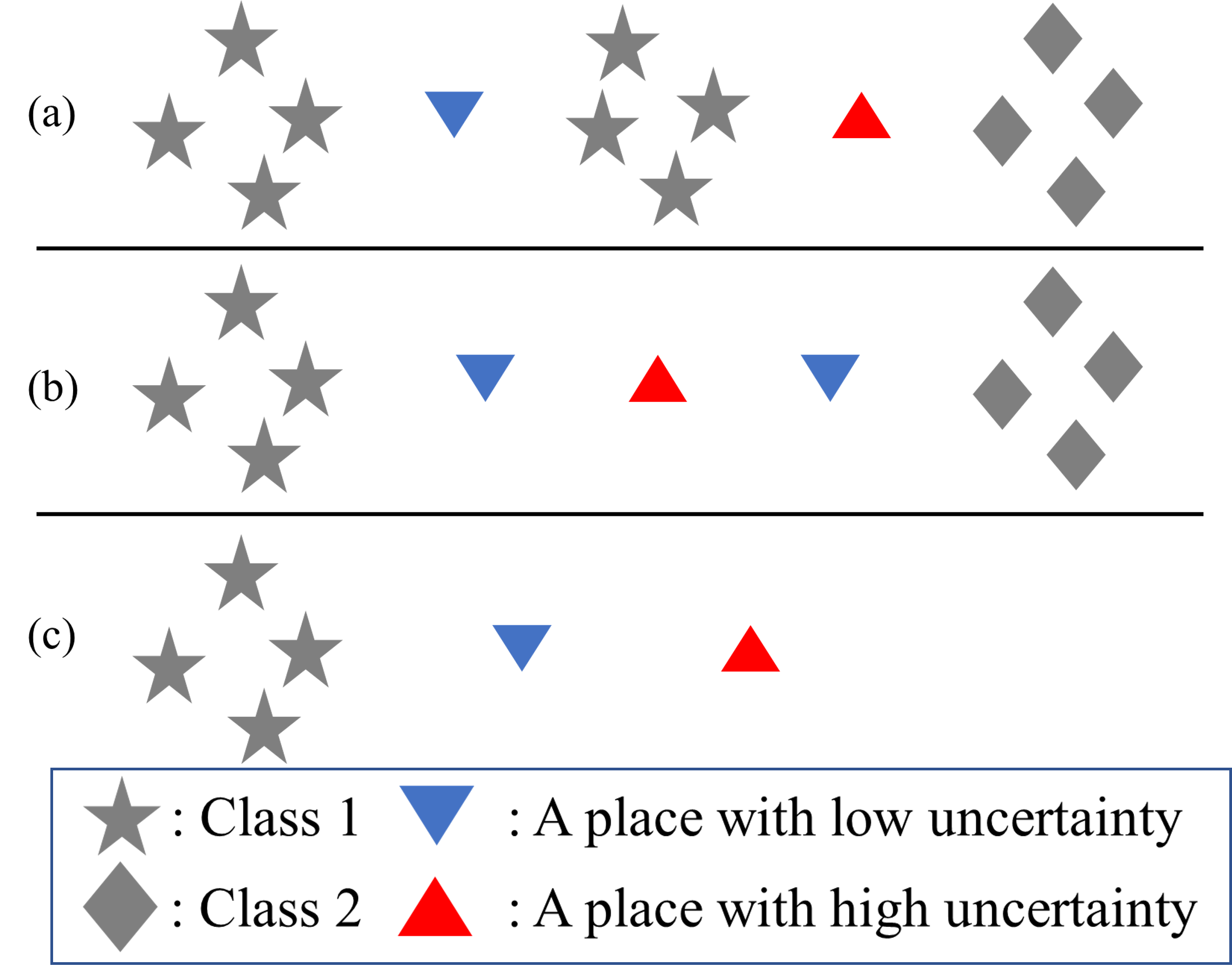} 
\caption{Uncertainty by location of data (Triangles with pointed top and bottom) with classes (Star/diamond-shaped shapes). (a) High uncertainty among different classes (A red triangle), (b) The closer data (Blue triangles) is to a particular class, the lower the uncertainty, (c) High uncertainty in data far away from existing data ranges (A red Triangle).}
\label{fig:1}
\end{figure}

\subsection{Uncertainty of classes}
\label{ch:uc_def}
Data among the same classes in the input space, denoted as the blue triangle in Figure \ref{fig:1} (a), tends to be easier to predict than data between different classes, corresponding to the red triangle in Figure \ref{fig:1} (a). The more diverse data in the neighborhood, the higher the uncertainty. Considering the distances from labelled data, the location close to a particular class, as shown in Figure \ref{fig:1} (b), has a low uncertainty. In sum, uncertainty should be estimated as the diversity of surrounding classes based on distance weights. Also, the further away from existing data ranges, the higher the uncertainty has to be, as shown in Figure \ref{fig:1} (c). The newly proposed metric for uncertainty in this paper captures the three characteristics of uncertainty depicted in Figure \ref{fig:1}.

\subsection{Distance-weighted class impurity (DCI)}
\label{ch:DCI} 

In this section we present the Distance-weighted Class Impurity, which represents the data uncertainty described in Section \ref{ch:uc_def}. 
We quantify the uncertainty of new data using not only class impurity, but also distances to labelled data.
The $K$ nearest labelled data points are selected by sorting the distances between labelled data and the new instance $x_i$ on the canonically normalized feature space.
Let $d_{ik}$ be the distance from $x_i$ to the $k$-th nearest labelled data where the index $k$ can run from 1 to $K$ \citep{Gou12}.
We define $d_{ik,\alpha}$ by exponentiating $d_{ik}$ by $\alpha$, 
then adding a small value $\epsilon$, as given in Equation \ref{eq:dika}. 
The parameter $\alpha$ controls the contribution of distances to uncertainty where the larger $\alpha$ corresponds to more enhanced differences between distances. The small value $\epsilon$ allows it to be used as a denominator.

\begin{equation}
d_{ik,\alpha} = d_{ik}^{\alpha} + \epsilon
\label{eq:dika}
\end{equation}

Let $y_{ik}$ be the label of the k-th nearest labelled data from $x_i$. 
We define Distance-weighted Class Impurity (DCI) for $x_i$ with three parameters, namely $\alpha$, $\beta$ and $K$, as shown in Equation \ref{eq2} where $\alpha$,$\beta$ are positive real numbers close to $1$.
\begin{equation}
{DCI}_i = \min_j 
\frac{\sum_{y_{ik} \notin Class_j, k \leq K}
\frac{1}{d_{ik,\alpha}}}
{\sum_{k=1}^{K} \frac{1}{d_{ik,\alpha}^{\beta}}} 
\label{eq2}
\end{equation} 

For $K$ labelled data nearest to $x_i$, the impurity of a particular class $j$ is calculated by weighting the reciprocal of $d_{ik,\alpha}$, 
and divide it by the sum of the reciprocals of $d_{ik,\alpha}^{\beta}$ with $\beta$.  
The DCI is defined as the minimum value of the impurity over all possible classes $j$.
For $\beta$, values greater than 1 are recommended to make DCI increase as the distance from the data points increases. Let us consider a case that the $K$ nearest points are equally distributed from $x_i$, i.e., $d_{I,k} = d$. The DCI in the case is $\min_j \frac{K-n_j}{K} d^{\alpha \cdot (\beta -1)}$ where $n_j$ is the number of data points in the j-th class. DCI becomes independent on distance $d$ when $\beta=1$, and becomes an increasing function of the $d$ when $ \beta > 1$. The higher $\beta$, the faster DCI increases with respect to the distance.

\begin{figure}
\centering
\includegraphics[scale=0.3]{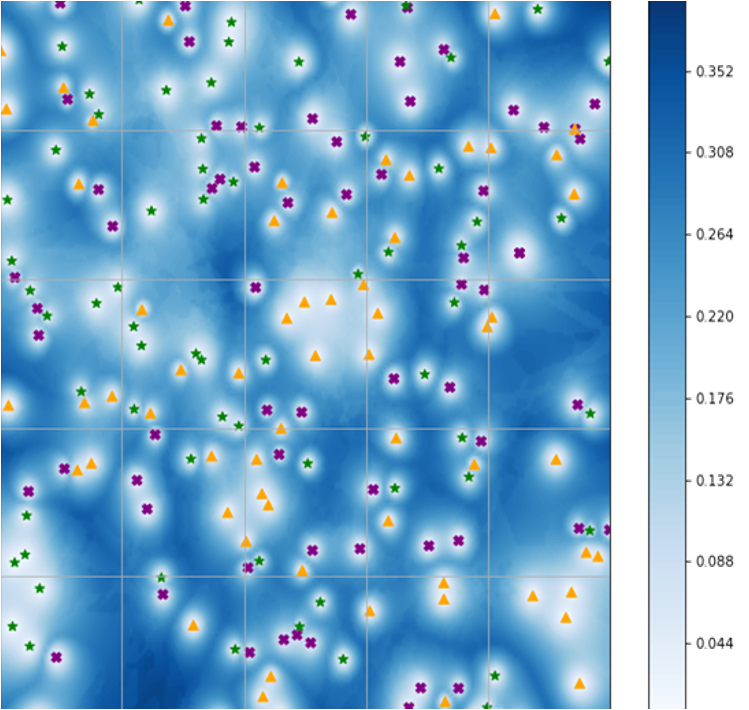} 
\caption{Distribution of Distance-weighted Class Impurities with Euclidean distances on 2-dimensional data. ($\alpha =1.5, \beta =1.2,$ and $K=20$)}
\label{fig:2}
\end{figure}

\subsection{Examples of 2-dimensional data} 
\label{ch:ex2d}
In this section, we devised randomly distributed data of three classes on a two-dimensional feature space and demonstrated DCIs. 
Figure \ref{fig:2} shows an example of a distribution of DCIs on the two-dimensional space. 
One can note that DCIs are low between the same classes and high between different classes. 

\begin{figure}
\centering
\includegraphics[scale=0.16]{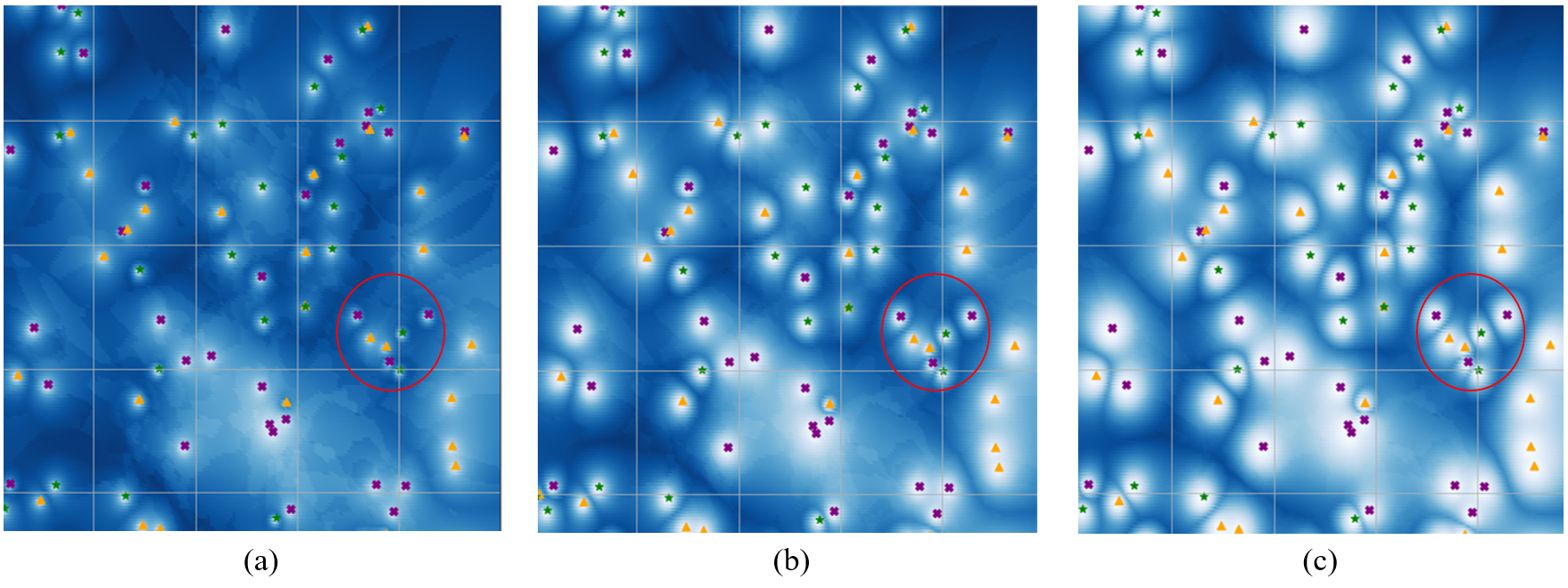} 
\caption{Variation of DCI distribution according to parameter $\alpha$. ($\beta =1.2$, $K=20$)  (a) $\alpha =1.0$, (b) $\alpha=1.5$, (c) $\alpha =2.0$.}
\label{fig:3}
\end{figure}

\begin{figure}
\centering
\includegraphics[scale=0.16]{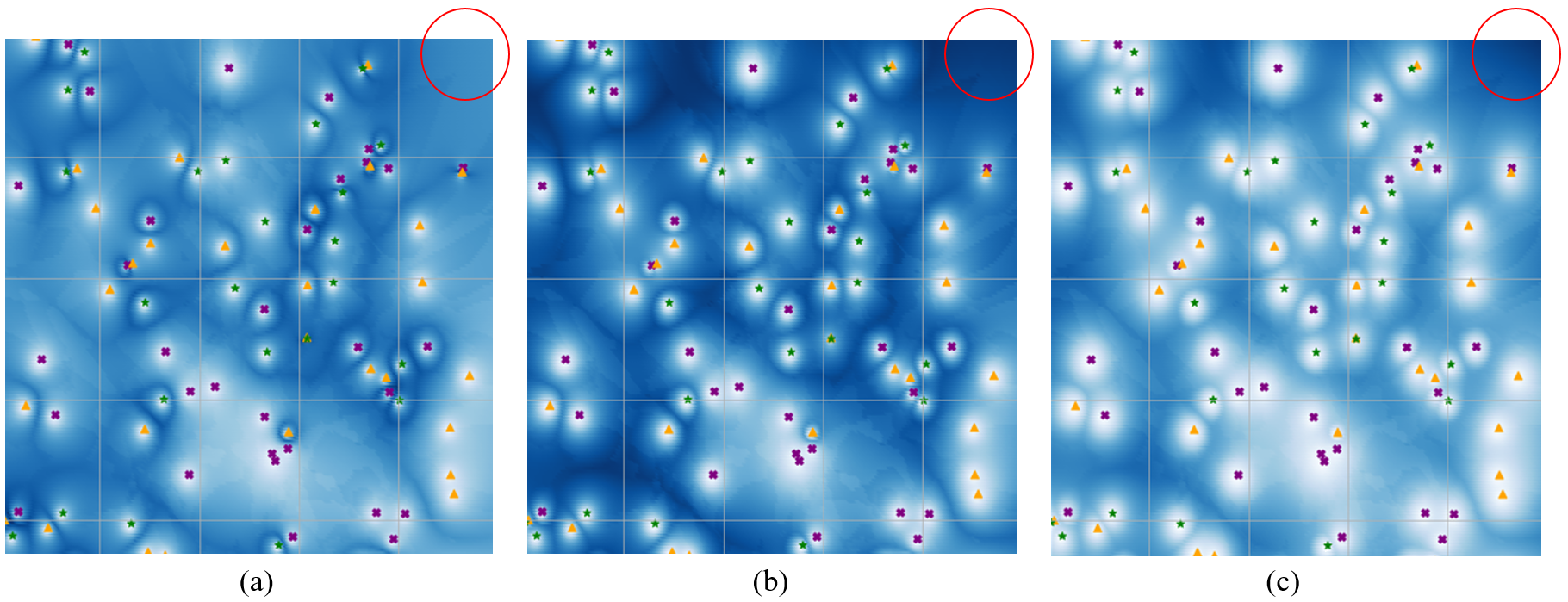} 
\caption{Variation of DCI distribution according to parameter $\beta$. ($\alpha =1.5$, $K=20$)  (a) $\beta =0.9$, (b) $\beta =1.1$, (c) $\beta =1.3$.}
\label{fig:4}
\end{figure}

Fig \ref{fig:3} and Fig \ref{fig:4} represents different distributions of DCI depending on parameters $\alpha$ and $\beta$.
The variation in parameter $\alpha$ results in different values of DCIs between distinct classes.
Figure \ref{fig:3} shows that boundaries between classes become clearer as $\alpha$ increases. 
The variation in parameter $\beta$ makes the DCI distribution different when the distance away from data. 
Figure \ref{fig:4} shows that as $\beta$ increases, DCIs grow in areas where data do not exist.

\section{EXPERIMENTS}
\label{exp} 

We tested the performance of DCIs with active learning tasks on three datasets, including UCI's Adult, Wine quality \citep{UCI} and MNIST dataset \citep{MNIST}. 
XGBoost \citep{XGB}, Random forest \citep{RF}, and CNN are implemented as the prediction models. 

Active learning experiments are conducted in the following order. Firstly, we trained the initial prediction models with a small number of randomly selected train data. 
And then, we estimated uncertainties for another five randomly selected unseen data, and added one datum with the highest uncertainty to the train set. 
After repeating processes of adding data several times, prediction models were updated. 
Each time the model was updated, prediction models were evaluated and compared.

\subsection{Adult dataset} 
\label{ch:Adult}
We utilized Adult dataset \citep{UCI} which is aimed to predict whether income is high or not. 
We removed the missing data, normalized integer, and continuous data as standard normal distributions, and encoded categorical data into one-hot vectors. 
The number of training and testing data is 30560, 15060, respectively. 

\subsubsection{Prediction model}
\label{ch:xgb_model}
We used the XGBoost ensemble model for binary classification. 
We used 5-fold cross validation twice with different random seeds. 
We evaluated on the test set in terms of the area under the receiver operating characteristic curve(AUROC) using average values of 10 XGBoost models.
We estimated uncertainty using the results of each XGBoost model in probability form. As shown in Equation \ref{eq3}, we averaged a difference of 0.5 from each probability and multiplied by -1.
When all probabilities are 0.5, the metric has the maximum value of 0 as it should be. 

\begin{equation}
Uncertainty_{XGB} = - \frac{1}{10} \sum_{i=1}^{10} | 0.5 - f_i (x) | 
\label{eq3} 
\end{equation}

\subsubsection{Results}
The initial XGBoost model was trained with 1,162 data. Data were added in four ways: random selection, DCI (High value) selection ($K$=20, $\alpha$=1.5, $\beta$=1.2), high XGBoost uncertainty selection (in Section \ref{ch:xgb_model}), and counter DCI (Low value) selection. 
The random selection method randomly selects data points to be added without the uncertainty evaluation. 
The high XGBoost uncertainty selection method evaluates XGBoost uncertainties for 5 random data points then selects the data point with the highest uncertainty. 
The DCI selection method evaluates DCIs for 5 random data points then selects the data point with the highest DCI. 
The counter DCI method is opposite to the DCI method selecting the data point with the lowest DCI instead of the highest DCI.
After adding 200 data in each method, the process of updating the model was repeated 29 times. 

\begin{figure}
\centering
\includegraphics[scale=0.23]{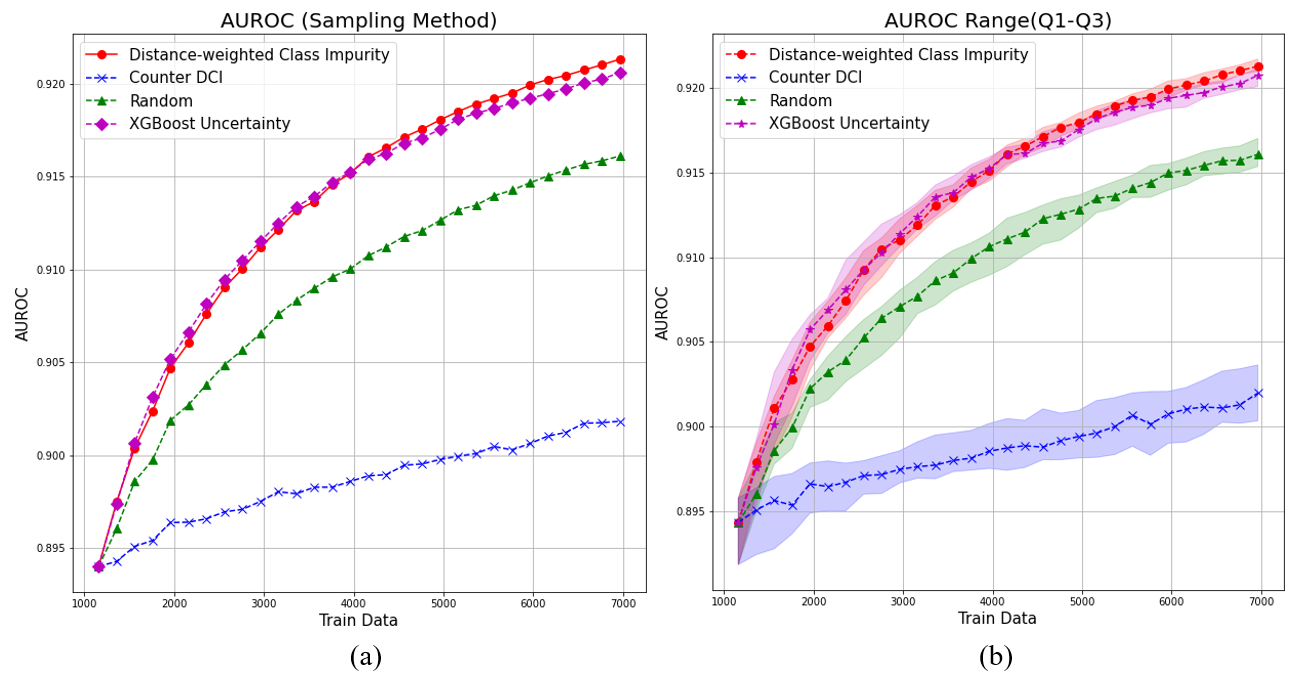} 
\caption{Active learning result on Adult dataset. (AUROCs) (a) Averages of 20 experiments, (b) Interquartile ranges and median values of 20 experiments.}
\label{fig:5}
\end{figure}

Figure \ref{fig:5} illustrates the results of the average AUROC by repeating the experiment 20 times. We found that the results of active learning through DCIs showed equal or slightly better performance to those using the uncertainty of XGBoost.

\subsubsection{Uncertainties vs. prediction accuracies}
\label{ch:split}
In this section, we compared uncertainties and prediction accuracies for adult dataset. 
XGBoost models were trained with 10, 15, 20, and 50 data. 
The uncertainty was calculated in two ways: the XGBoost uncertainty (in Section \ref{ch:xgb_model}) and the DCI.
DCIs were calculated for test data with various $\alpha$s and $\beta$s. 
After calculating uncertainties, we sorted test data by each uncertainty and evaluated accuracies of XGBoost models every 10 percentile.

\begin{figure}
\centering
\includegraphics[scale=0.19]{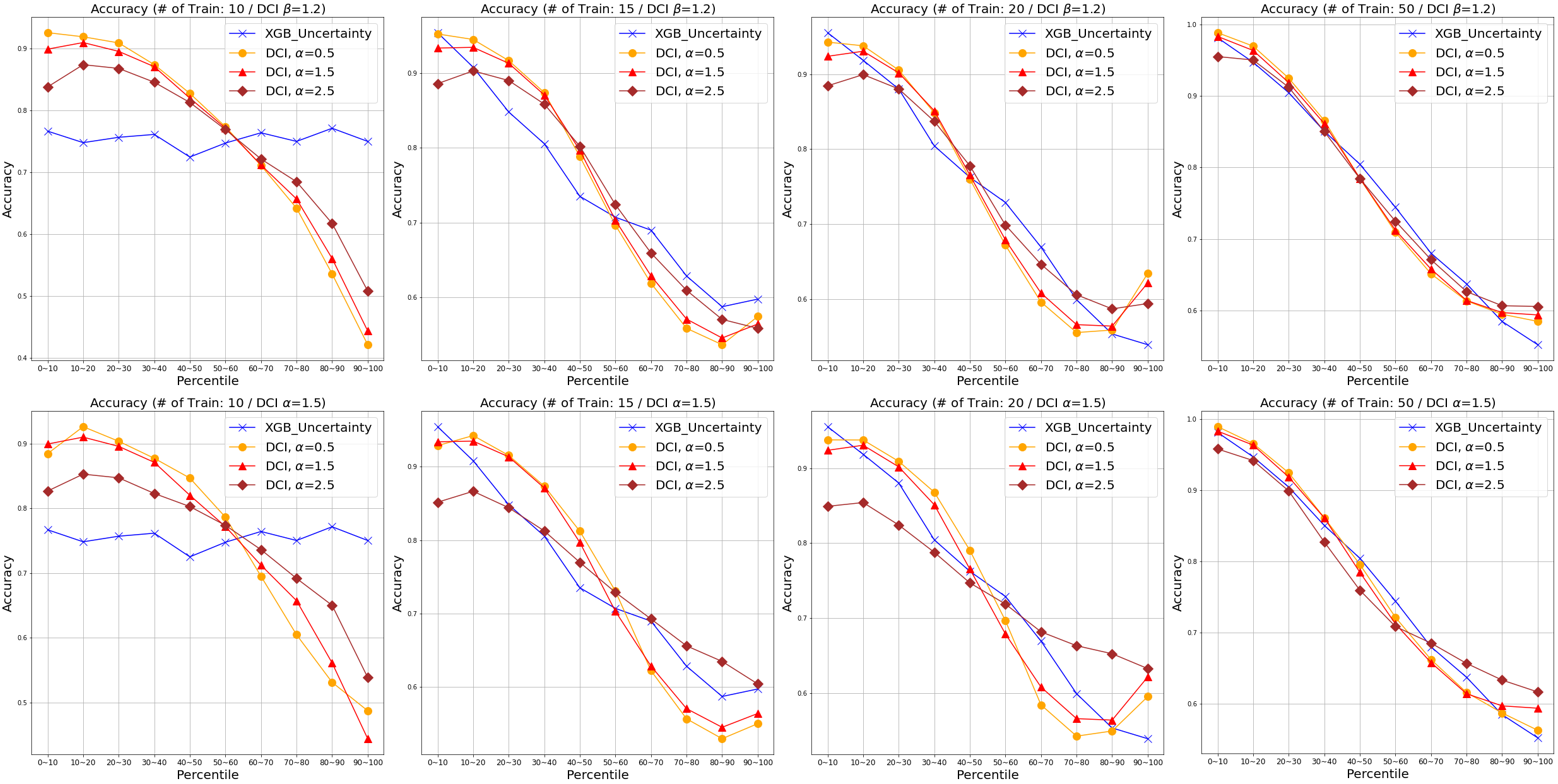} 
\caption{Uncertainties vs. prediction accuracies. (Averages of 20 experiments)
Each graph represents accuracies of XGBoost models every 10 percentile.
The upper graphs contain the results for DCIs with various $\alpha$s and the lower graphs contain the results for DCIs with various $\beta$s.
From the left, each graph shows prediction accuracies of XGBoost models trained with 10, 15, 20, and 50 data.}
\label{fig:new}
\end{figure}

Figure \ref{fig:new} illustrates the results of prediction accuracies with uncertainty levels. Accuracies are averaged over 20 random training splits.
Parameter $\alpha$ and $\beta$ have no noticeable effect on the accuracy. 
When data are collected at a certain level or more, the change in accuracy is similar regardless of the type of uncertainty, as shown in right side of Figure \ref{fig:new}.
But when data are not collected at a certain level, accuracy differences occur only depending on the DCI level, as shown in the left side of Figure \ref{fig:new}.

\subsection{Wine quality dataset} 
\label{ch:Wine}
We utilized Wine quality dataset \citep{UCI} to predict wine quality by regression. It consisted of 1,599 red wine data and 4,898 white wine data. 
The response variable is ordinal, however we regarded it as regression, known as ordinal regression. 
Experiments were done on white and red wines, separately. 
As test sets, 599 red wines, and 1,398 white wines were used.
All features were canonically normalized. 
The response variable, wine quality, was treated as categorical data to evaluate DCIs.

\subsubsection{Prediction model}
\label{ch:rf_model}
We employed the random forest model consisted of 100 decision trees for regression. 
The uncertainty based on random forest models were estimated by the standard deviation of each decision tree’s prediction \citep{lie}.
Random forest models were evaluated on the test set in terms of the root mean square error (RMSE). 

\subsubsection{Results}
The initial random forest models were learned with 200 data for red wine and 500 data for white wine. Data were added in four ways: random selection, DCI (High Value) selection ($K=20$, $\alpha=1.5$, $\beta=1.2$), 
high random forest uncertainty selection (in Section \ref{ch:rf_model}), and counter DCI (Low value) selection. 
After adding 10 (resp. 25) for red (resp. white) wine data in each method, the process of updating the model was repeated 18 (resp. 24) times for red (resp. white) wines. 

\begin{figure}
\centering
\includegraphics[scale=0.23]{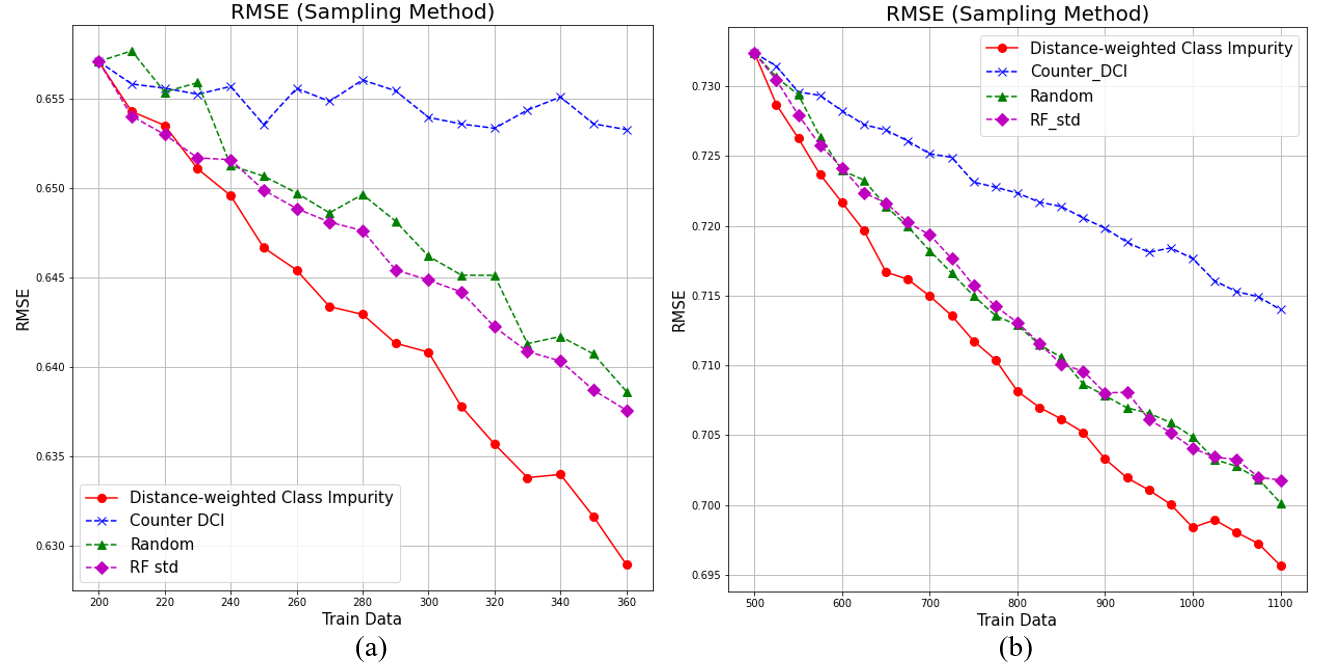} 
\caption{Active learning result on Wine quality dataset. (RMSEs) (a) Averages of 30 red wine experiments, (b) Averages of 30 white wine experiments.}
\label{fig:6}
\end{figure}

Figure \ref{fig:6} presents the number of train data versus average RMSEs where the RMSE is averaged over 30 random training and test splits. (a) is the result of red wine data, and (b) is the result of white wine data. 
We found that the DCI selection method outperformed in active learning tasks to other methods including uncertainty of random forest.

\subsection{MNIST dataset}
\label{ch:MNIST}
We used MNIST of which goal is to classify 10 digits. DCIs were calculated in two different feature space. The first method used Euclidean distance between 784($28\times28$) of original features, and the second method used principal component analysis(PCA). Classification tasks were conducted via CNNs using $28\times28$ pixel grayscale images.

\subsubsection{Prediction model}
\label{ch:cnn_model}

\begin{figure}
\centering
\includegraphics[scale=0.3]{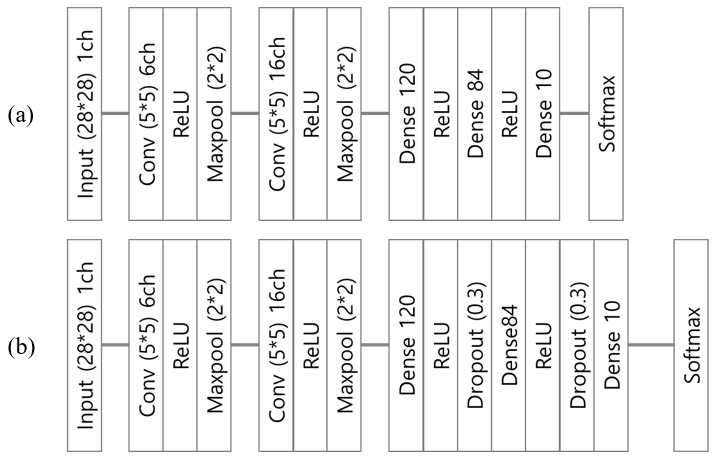} 
\caption{CNN architectures on classifying MNIST tasks. (a) Simple, (b) Bayesian.}
\label{fig:7}
\end{figure}

We used a simple CNN and a Bayesian CNN \citep{Gal_baye} for uncertainty estimations, as shown in Figure \ref{fig:7}. For the Bayesian CNN, uncertainty can be estimated by mean standard deviation \citep{Kampffmeyer, Shim}, where the mean denotes average over all possible classes. The standard deviations of the probabilities for each class were averaged for 10 of CNN results. For simple CNNs, uncertainty was estimated as $1-(probability\ of\ predicted\ digit)$ \citep{Sett}.

\subsubsection{Results of simple CNNs}
\label{ch:simple}
The initial CNN models were trained with 10 of data. Data were added in six ways: 
random selection, DCI(High Value) selection($K=10$, $\alpha=1.5$, $\beta=1.2$), 
counter DCI(Low Value) selection, high simple CNN uncertainty selection(in Section \ref{ch:cnn_model}), high DCI for 10 principal components (PCs) selection, and counter DCI for 10 PCs. Each PC was multiplied by its explained variance ratio and then used. 
As for the other datasets, the selection methods based on highest (resp. lowest) DCI measure DCIs of 5 randomly chosen data points then select the data point with the highest (resp. lowest) DCI to be added into the training set.
After adding 5 data in each method, the process of updating the model was repeated 18 times. 

\begin{figure}
\centering
\includegraphics[scale=0.23]{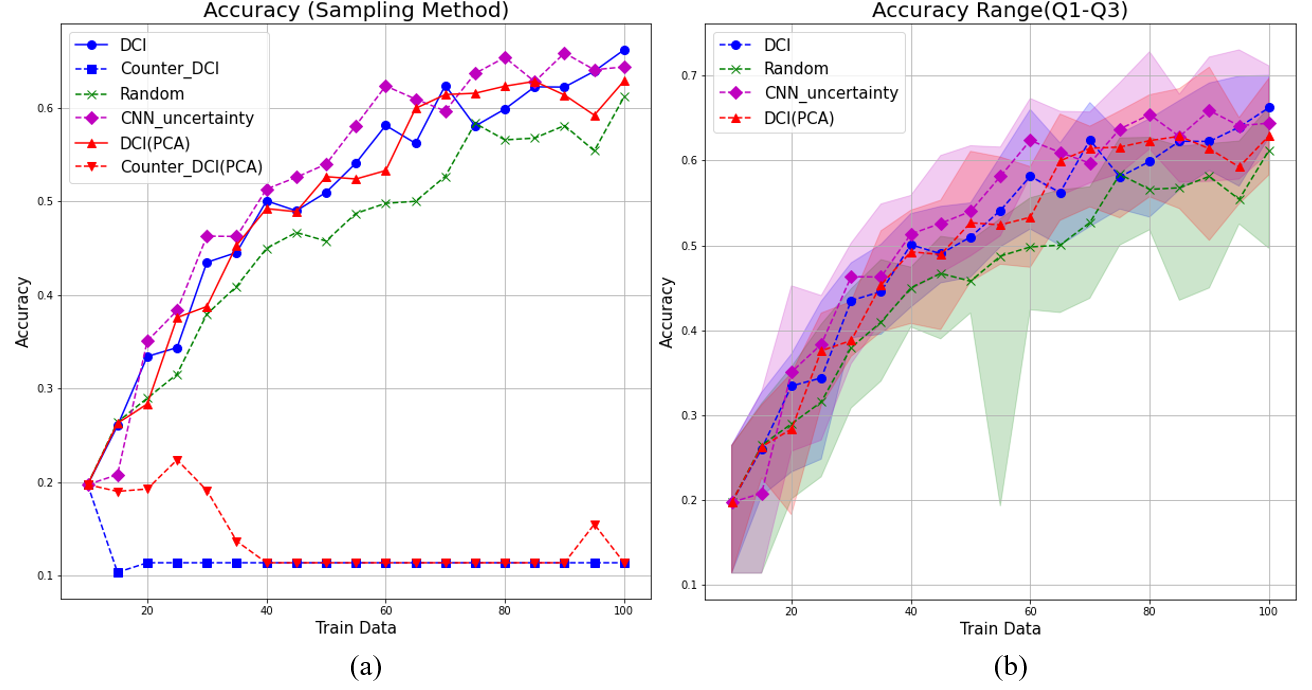} 
\caption{Active learning result on MNIST dataset using simple CNNs. (a) Averages of 30 experiments, (b) Interquartile ranges and median values of 30 experiments.}
\label{fig:8}
\end{figure}

Figure \ref{fig:8} shows the results of the average accuracy over 30 random training and test splits. The DCI showed equal or slightly lower performance as using CNN uncertainty.

\begin{figure}
\centering
\includegraphics[scale=0.27]{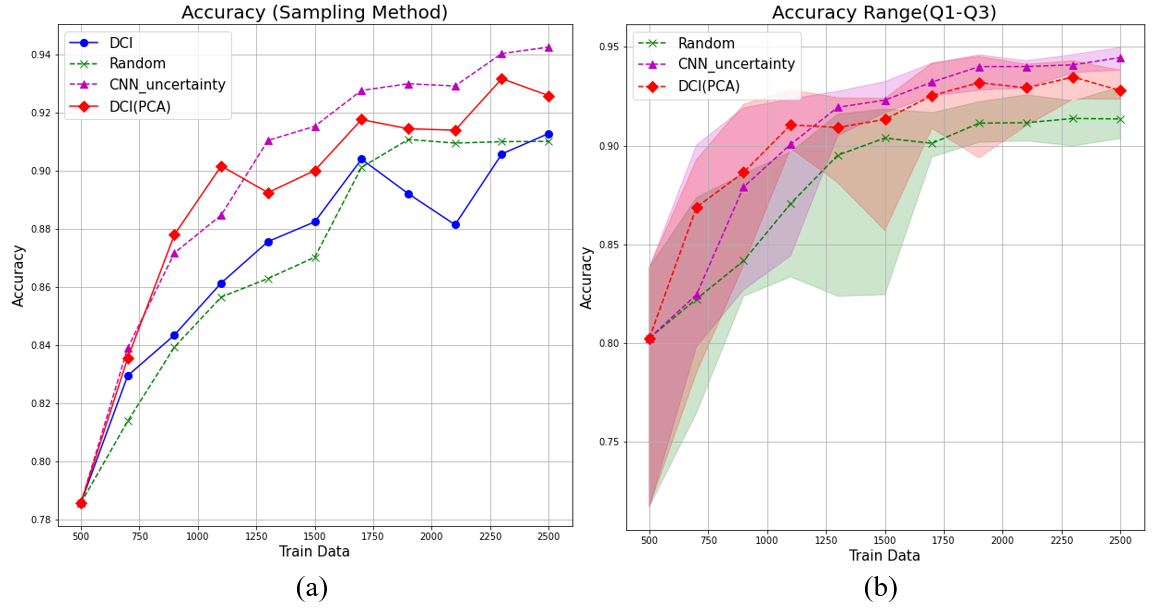} 
\caption{Active learning result on MNIST dataset using Bayesian CNNs. (a) Averages of 20 experiments, (b) Interquartile ranges and median values of 20 experiments.}
\label{fig:9}
\end{figure}

\subsubsection{Results of Bayesian CNNs}
\label{ch:bayes}
Initial CNN models were trained on the original features and on the PCA reduced features, respectively, 500 data. 
Data were added in four ways: random selection, DCI (High Value) selection ($K=20$, 
$\alpha=1.5$, $\beta=1.2$), high Bayesian CNN Uncertainty selection (in Section \ref{ch:cnn_model}), and DCI for 20 PCs (High value) selection. 
Each PC was multiplied by its explained variance ratio and then used. 
After adding 200 data in each method, the process of updating the model including PCAs was repeated 10 times.

Figure \ref{fig:9} shows the results of the average accuracy over 20 random training and test splits. 
DCIs with PCA features showed similar performance to those using uncertainty of Bayesian CNNs.
We here would like to re-emphasize that DCI does not require building and updating classification models, thus more time-efficient than CNN uncertainty.

\subsection{Discussion}
\label{discuss}
We verified the performance of the DCI on various datasets and machine learning algorithms. 
In Section \ref{ch:Wine}, DCIs effectively sampled on not only the classification task but also the ordinal regression task. 
DCI was also applicable even when discrete features were included, as one-hot encoding, as shown in Section \ref{ch:Adult}. 
Moreover, DCIs worked effectively in the early stages of data acquisition, as shown in Section \ref{ch:split}. 
We observed that there were performance differences depending on the feature space for high-dimensional data such as images, as shown in Section \ref{ch:MNIST}.

\section{CONCLUSION}

Effective sampling is critical in the process of data acquisition for machine learning. In this study, we proposed a quantitative metric for uncertainty called the Distance-weighted Class Impurity without specifying machine learning models. 
We verified the performance of the DCI on various active learning tasks. 
The DCI enables collecting data to be effective regardless of machine learning model. 
It significantly reduced the time for building and updating machine learning models. 
We believe that even those who lack understanding of machine learning can achieve effective data acquisition by deploying the DCI.

\bibliographystyle{plainnat}
\bibliography{0_bib}

\begin{thebibliography}{18}
\providecommand{\natexlab}[1]{#1}
\providecommand{\url}[1]{\texttt{#1}}
\expandafter\ifx\csname urlstyle\endcsname\relax
  \providecommand{\doi}[1]{doi: #1}\else
  \providecommand{\doi}{doi: \begingroup \urlstyle{rm}\Url}\fi

\bibitem[Breiman(2001)]{RF}
Leo Breiman.
\newblock Random forests.
\newblock \emph{Machine Learning}, 45\penalty0 (1):\penalty0 5--32, 2001.
\newblock ISSN 0885-6125.
\newblock \doi{10.1023/A:1010933404324}.

\bibitem[Chen and Guestrin(2016)]{XGB}
Tianqi Chen and Carlos Guestrin.
\newblock Xgboost: A scalable tree boosting system.
\newblock In \emph{Proceedings of the 22nd ACM SIGKDD International Conference
  on Knowledge Discovery and Data Mining}, KDD '16, page 785–794, New York,
  NY, USA, 2016. Association for Computing Machinery.
\newblock ISBN 9781450342322.
\newblock \doi{10.1145/2939672.2939785}.
\newblock URL \url{https://doi.org/10.1145/2939672.2939785}.

\bibitem[Cohn et~al.(1996)Cohn, Ghahramani, and Jordan]{Cohn}
David~A. Cohn, Zoubin Ghahramani, and Michael~I. Jordan.
\newblock Active learning with statistical models.
\newblock 4\penalty0 (1):\penalty0 129–145, March 1996.
\newblock ISSN 1076-9757.

\bibitem[Dua and Graff(2017)]{UCI}
Dheeru Dua and Casey Graff.
\newblock {UCI} machine learning repository, 2017.
\newblock URL \url{http://archive.ics.uci.edu/ml}.

\bibitem[Gal and Ghahramani(2016)]{Gal_baye}
Yarin Gal and Zoubin Ghahramani.
\newblock Dropout as a bayesian approximation: Representing model uncertainty
  in deep learning.
\newblock In Maria~Florina Balcan and Kilian~Q. Weinberger, editors,
  \emph{Proceedings of The 33rd International Conference on Machine Learning},
  volume~48 of \emph{Proceedings of Machine Learning Research}, pages
  1050--1059, New York, New York, USA, 20--22 Jun 2016. PMLR.

\bibitem[Gal et~al.(2017)Gal, Islam, and Ghahramani]{Gal_image}
Yarin Gal, Riashat Islam, and Zoubin Ghahramani.
\newblock Deep {B}ayesian active learning with image data.
\newblock In Doina Precup and Yee~Whye Teh, editors, \emph{Proceedings of the
  34th International Conference on Machine Learning}, volume~70 of
  \emph{Proceedings of Machine Learning Research}, pages 1183--1192. PMLR,
  06--11 Aug 2017.

\bibitem[Gou et~al.(2012)Gou, Du, Zhang, and Xiong]{Gou12}
Jianping Gou, Lan Du, Y.~Zhang, and Taisong Xiong.
\newblock A new distance-weighted k-nearest neighbor classifier.
\newblock \emph{The Journal of Information and Computational Science},
  9:\penalty0 1429--1436, 2012.

\bibitem[Kampffmeyer et~al.(2016)Kampffmeyer, Salberg, and
  Jenssen]{Kampffmeyer}
Michael Kampffmeyer, Arnt-Børre Salberg, and Robert Jenssen.
\newblock Semantic segmentation of small objects and modeling of uncertainty in
  urban remote sensing images using deep convolutional neural networks.
\newblock 07 2016.
\newblock \doi{10.1109/CVPRW.2016.90}.

\bibitem[LeCun and Cortes(2010)]{MNIST}
Yann LeCun and Corinna Cortes.
\newblock {MNIST} handwritten digit database.
\newblock 2010.
\newblock URL \url{http://yann.lecun.com/exdb/mnist/}.

\bibitem[Ließ et~al.(2012)Ließ, Glaser, and Huwe]{lie}
Mareike Ließ, Bruno Glaser, and Bernd Huwe.
\newblock Uncertainty in the spatial prediction of soil texture: Comparison of
  regression tree and random forest models.
\newblock \emph{Geoderma}, 170:\penalty0 70--79, 2012.
\newblock ISSN 0016-7061.
\newblock \doi{https://doi.org/10.1016/j.geoderma.2011.10.010}.
\newblock URL
  \url{https://www.sciencedirect.com/science/article/pii/S0016706111002953}.

\bibitem[McCallum and Nigam(1998)]{McCallum}
Andrew McCallum and Kamal Nigam.
\newblock Employing em and pool-based active learning for text classification.
\newblock In \emph{Proceedings of the Fifteenth International Conference on
  Machine Learning}, ICML '98, page 350–358, San Francisco, CA, USA, 1998.
  Morgan Kaufmann Publishers Inc.
\newblock ISBN 1558605568.

\bibitem[{Niaf} et~al.(2011){Niaf}, {Flamary}, {Lartizien}, and {Canu}]{Niaf}
É. {Niaf}, R.~{Flamary}, C.~{Lartizien}, and S.~{Canu}.
\newblock Handling uncertainties in svm classification.
\newblock In \emph{2011 IEEE Statistical Signal Processing Workshop (SSP)},
  pages 757--760, 2011.
\newblock \doi{10.1109/SSP.2011.5967814}.

\bibitem[{Patra} and {Bruzzone}(2012)]{Patra}
S.~{Patra} and L.~{Bruzzone}.
\newblock A batch-mode active learning technique based on multiple uncertainty
  for svm classifier.
\newblock \emph{IEEE Geoscience and Remote Sensing Letters}, 9\penalty0
  (3):\penalty0 497--501, 2012.
\newblock \doi{10.1109/LGRS.2011.2172770}.

\bibitem[Platt(1999)]{Platt99}
John~C. Platt.
\newblock Probabilistic outputs for support vector machines and comparisons to
  regularized likelihood methods.
\newblock In \emph{ADVANCES IN LARGE MARGIN CLASSIFIERS}, pages 61--74. MIT
  Press, 1999.

\bibitem[Settles and Craven(2008)]{Sett}
Burr Settles and Mark Craven.
\newblock An analysis of active learning strategies for sequence labeling
  tasks.
\newblock In \emph{Proceedings of the Conference on Empirical Methods in
  Natural Language Processing}, EMNLP '08, page 1070–1079, USA, 2008.
  Association for Computational Linguistics.

\bibitem[Shen et~al.(2018)Shen, Yun, Lipton, Kronrod, and Anandkumar]{shen}
Yanyao Shen, Hyokun Yun, Zachary~C. Lipton, Yakov Kronrod, and Animashree
  Anandkumar.
\newblock Deep active learning for named entity recognition, 2018.

\bibitem[{Shim} et~al.(2020){Shim}, {Kang}, and {Cho}]{Shim}
J.~{Shim}, S.~{Kang}, and S.~{Cho}.
\newblock Active learning of convolutional neural network for cost-effective
  wafer map pattern classification.
\newblock \emph{IEEE Transactions on Semiconductor Manufacturing}, 33\penalty0
  (2):\penalty0 258--266, 2020.
\newblock \doi{10.1109/TSM.2020.2974867}.

\bibitem[{Wang} et~al.(2017){Wang}, {Zhang}, {Li}, {Zhang}, and {Lin}]{Wang}
K.~{Wang}, D.~{Zhang}, Y.~{Li}, R.~{Zhang}, and L.~{Lin}.
\newblock Cost-effective active learning for deep image classification.
\newblock \emph{IEEE Transactions on Circuits and Systems for Video
  Technology}, 27\penalty0 (12):\penalty0 2591--2600, 2017.
\newblock \doi{10.1109/TCSVT.2016.2589879}.

\end{thebibliography}
\end{document}